\pdfoutput=1
\documentclass[letterpaper, 10 pt, conference]{ieeeconf}  

\IEEEoverridecommandlockouts                              

\usepackage{graphics} 
\usepackage{epsfig} 
\usepackage{mathptmx} 
\usepackage{times} 
\usepackage{amsmath} 
\usepackage{amssymb}  
\usepackage{comment}
\usepackage{tabularx, booktabs, array}
\usepackage{todonotes}

\usepackage{xcolor,colortbl}
\usepackage[binary-units=true]{siunitx}
\usepackage{bm}
\usepackage{rotating} 
\usepackage{multirow}

\usepackage[outline]{contour}

\usepackage{cite}

\usepackage{subcaption}

\newcommand{\reffig}[1]{Fig.~\ref{#1}}
\newcommand{\reftab}[1]{Tab.~\ref{#1}}
\newcommand{\refsec}[1]{Sec.~\ref{#1}}

\newcommand{\etal}{et al.~}

\newcommand{\PreserveBackslash}[1]{\let\temp=\\#1\let\\=\temp}
\newcolumntype{C}[1]{>{\PreserveBackslash\centering}p{#1}}
\newcolumntype{R}[1]{>{\PreserveBackslash\raggedleft}p{#1}}
\newcolumntype{L}[1]{>{\PreserveBackslash\raggedright}p{#1}}
\newcommand\ph{$\phantom{1}$} 

\usepackage[outline]{contour}

\usepackage{tikz, tikz-3dplot}
\usetikzlibrary{calc,arrows,arrows.meta,backgrounds,tikzmark,shapes.geometric, decorations.pathreplacing,decorations.markings, spy}

\definecolor{ph-purple}{RGB}{129, 39, 232}
\definecolor{ph-blue}{RGB}{5, 131, 227}
\definecolor{ph-dark-blue}{RGB}{0, 91, 187}
\definecolor{ph-gray}{rgb}{0.5, 0.5, 0.5}
\definecolor{ph-orange}{RGB}{227, 127, 5}
\definecolor{ph-green}{RGB}{0, 135, 124}
\definecolor{ph-yellow}{RGB}{ 222, 194, 11}
\definecolor{ph-light-green}{RGB}{181, 209, 21}
\definecolor{ph-red}{RGB}{250, 101, 60}
\definecolor{ph-dark-red}{RGB}{210, 61, 20}
\definecolor{Gainsboro}{RGB}{220, 220, 220}
\colorlet{ph-orange-light}{ph-orange!70}
\colorlet{ph-blue-light}{ph-blue!70}
\colorlet{ph-purple-light}{ph-purple!70}
\colorlet{ph-green-light}{ph-green!70}
\definecolor{ph-light-gray}{rgb}{0.75, 0.75, 0.75}
\definecolor{road}{RGB}{	255, 0, 255	}
\definecolor{parking}{RGB}{	255, 150, 255	}
\definecolor{unlabeled}{RGB}{	0, 0, 0	}

\tikzset{
	on each segment/.style={
		decorate,
		decoration={
			show path construction,
			moveto code={},
			lineto code={
				\path [#1]
				(\tikzinputsegmentfirst) -- (\tikzinputsegmentlast);
			},
			curveto code={
				\path [#1] (\tikzinputsegmentfirst)
				.. controls
				(\tikzinputsegmentsupporta) and (\tikzinputsegmentsupportb)
				..
				(\tikzinputsegmentlast);
			},
			closepath code={
				\path [#1]
				(\tikzinputsegmentfirst) -- (\tikzinputsegmentlast);
			},
		},
	},
	mid arrow/.style={postaction={decorate,decoration={
				markings,
				mark=at position .5 with {\arrow[#1]{stealth}}
	}}},
	pos arrow/.style 2 args={postaction={decorate,decoration={
				markings,
				mark=at position [#2] with {\arrow[#1]{stealth}}
	}}},
	custom arrow/.style={postaction={decorate,decoration={
			markings,
			mark=at position .528 with {\arrow[#1]{stealth}}
	}}},
	custom arrow2/.style={postaction={decorate,decoration={
			markings,
			mark=at position .475 with {\arrow[#1]{stealth}}
	}}},
}

\usepackage[draft,nomargin,inline]{fixme}
\fxsetup{theme=color}

\usepackage[hidelinks, bookmarks=false]{hyperref}
\usepackage{url}

\definecolor{car}{RGB}{245, 150, 100}
\definecolor{moving-car}{RGB}{255, 51, 0}

\title{\LARGE \bf
Abstract Flow for Temporal Semantic Segmentation on the Permutohedral Lattice
}

\author{Peer Sch\"utt, Radu Alexandru Rosu and Sven Behnke
\thanks{All authors are with Autonomous Intelligent Systems Group, Computer Science VI, University of Bonn, Germany, {\tt \{schuett, rosu, behnke\}@ais.uni-bonn.de.}}}%

\begin{document}

\maketitle
\thispagestyle{empty}
\pagestyle{empty}

\begin{abstract}


Semantic segmentation is a core ability required by autonomous agents, as being able to distinguish which parts of the scene belong to which object class is crucial for navigation and interaction with the environment. Approaches which use only one time-step of data cannot distinguish between moving objects nor can they benefit from temporal integration. In this work, we extend a backbone LatticeNet to process temporal point cloud data. Additionally, we take inspiration from optical flow methods and propose a new module called Abstract Flow which allows the network to match parts of the scene with similar abstract features and gather the information temporally. We obtain state-of-the-art results on the SemanticKITTI dataset that contains LiDAR scans from real urban environments. We share the PyTorch implementation of TemporalLatticeNet at \normalfont{\url{https://github.com/AIS-Bonn/temporal_latticenet}}.

\end{abstract}

\section{Introduction}

Semantic segmentation of 3D point clouds is the process of predicting a class for every point in the cloud. This is especially challenging for 3D point clouds, due to undersampling of the scene and a lack of explicit structure in the cloud.

Current approaches rely on projecting the 3D point cloud to 2D images \cite{shi2020spsequencenet,duerr2020lidar} or embed it into a dense volumetric grid \cite{tchapmi2017segcloud,rethage2018fully}. These approaches are, however, suboptimal since 2D images are not a natural representation for 3D point clouds while volumetric grids can be slow and memory intensive.
We propose to use the permutohedral lattice as an alternative representation to apply convolutions on a data structure that more closely resembles the input and can process full scans at interactive speeds.

In this work, we consider the point clouds as being recorded continuously by a sensor like a laser scanner or a depth camera. Processing individual point clouds ignores the temporal information, rendering the agent incapable of distinguishing between moving and stationary objects or integrating evidence over time~\cite{rosu2021latticenetauro,shi2020spsequencenet,duerr2020lidar}.


LatticeNet~\cite{rosu2019latticenet} is a efficient network based on the permutohedral lattice that has shown state-of-the-art results for semantic segmentation of point clouds. It has recently been extended to process temporal information~\cite{rosu2021latticenetauro}. However, the temporal fusion was performed only with simple recurrence using fully connected layers. 

We present TemporalLatticeNet, an extension to LatticeNet that utilizes recurrent processing by evaluating Long Short-Term Memory (LSTM)~\cite{hochreiter1997long} and Gated Recurrent Units (GRU)~\cite{cho2014properties} as modules for temporal fusion. Additionally, we propose a novel module called Abstract Flow (AFlow) which can gather temporal information by matching abstract features in lattice space.

We show competitive results on the temporal SemanticKITTI dataset~\cite{behley2019semantickitti}, but with a faster processing speed than other methods.

\begin{figure}[t]
\centering
\includegraphics[width = \linewidth]{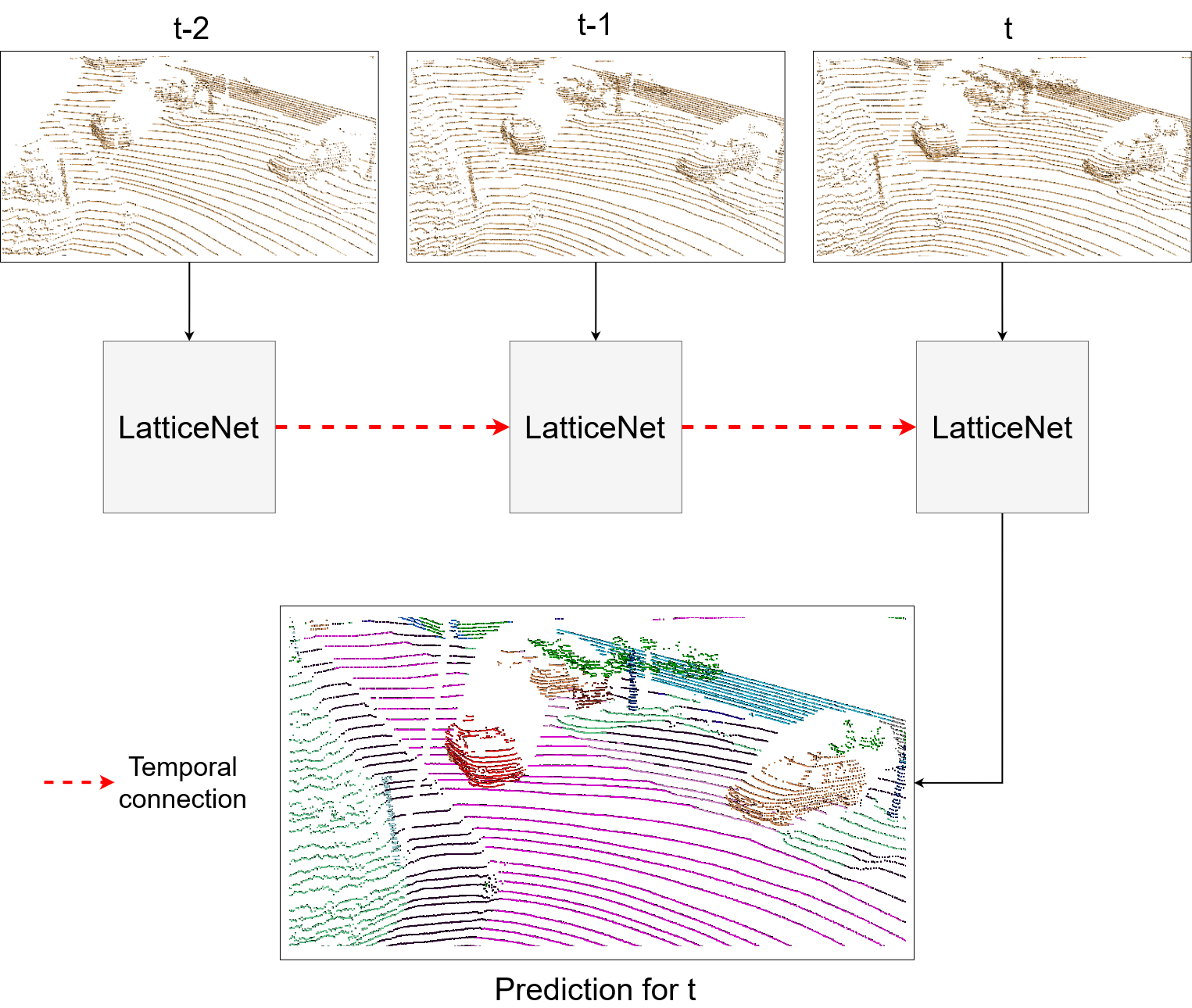}
\caption{We use multiple consecutive point clouds as input to our backbone network. The feature matrices of different timesteps are fused together to allow information propagation through time. The per-point semantic classes of the last point cloud in the sequence are predicted.}
\label{fig:overview_arch}
\end{figure}

Our contributions presented in this work include:
\begin{itemize}
	\item Extensions to the LatticeNet architecture in order to improve recurrent processing,
	\item Novel module inspired by optical flow methods adapted to the permutohedral lattice that improves the aggregation of temporal information, and
 	\item Competitive results on the SemanticKITTI dataset but with faster processing speed than other methods.

\end{itemize}

\section{Related Work}

In this section, temporal semantic segmentation will be discussed. 
For an overview on general semantic segmentation of point clouds without temporal information we refer to Rosu~\etal\cite{rosu2021latticenetauro}

Temporal semantic segmentation approaches can be divided in two types, depending on their input: i) methods which process sequences of clouds in a recurrent manner to predict the class labels and ii) methods that accumulate multiple clouds into one single cloud to solve the task as a single-frame segmentation. 

The advantage of the first type of methods is that at each time-step only the current point cloud needs to be processed as the past information has already been summarized and stored in memory. However, they require more complex architectures than the methods that na\"{\i}vely segment a large aggregated point cloud.

Shi~\etal\cite{shi2020spsequencenet} present SpSequenceNet, a U-Net based architecture for temporal semantic segmentation.
Two consecutive point clouds are voxelized and given as input to a shared encoder network. Vertical connection are added between the two timesteps in order to gather temporal information before decoding the representation into class probabilities of the last point cloud.
They designed two modules to combine the information from the two consecutive point clouds: the Cross-frame Global Attention~(CGA) and the Cross-frame Local Interpolation~(CLI) module. 
CGA is inspired by self-attention and allows the network to gather and fuse global information from the previous cloud. CLI allows to attend to local information by fusing the features from nearby points with a dynamic per-point weighting.
In contrast to our approach, Shi~\etal\cite{shi2020spsequencenet} can only process sequences of length two and they voxelize the point cloud, which leads to a loss of information and discretization artifacts.
Additionally, CLI per-point weighting is calculated using distances to k-nearest neighbors in 3D space, while our approach calculates the distance in the abstract feature space using the 1-hop neighborhood of a permutohedral lattice.

Duerr~\etal\cite{duerr2020lidar} present their recurrent architecture TemporalLidarSeg that uses temporal memory alignment to predict the semantic labels of sequences of point clouds. Their sequences have the potential of unlimited length. They project the 3D point clouds onto the 2D plane and use a U-net backbone network to output per-frame feature matrices. 
These feature matrices are then combined with the feature matrices of the hidden state using the temporal memory unit, which uses the real-world poses of each point cloud to compute the transformation from the coordinate system of the hidden to the current state. 
The 2D semantic segmentation is then projected back into the 3D representation. 
Similar to our approach, they require the poses of the point clouds, but they additionally need the mapping from 3D to 2D and therefore don't work directly on the point cloud. Additionally, our approach uses multiple fusion points in contrast to one in their approach.

Kernel Point Convolution (KPConv)~\cite{thomas2019KPConv} operates directly on the point clouds by assigning convolution weights to a set of kernel points located in Euclidean space. Points in the vicinity of these kernel points are weighted and summed together to feature vectors. The kernel function is defined as the correlation between the location of the kernel point and the distance to the points in the radius neighborhood. To be robust to varying densities, the input clouds are subsampled at every layer of the network using a grid subsampling and the radius neighborhood of the convolution is adapted. The kernel points are usually static, but can also be learned by the network itself to adapt to more challenging tasks. Due to memory limitations, their approach cannot process one full point cloud for outdoor scenes. Therefore, Thomas~\etal fit multiple overlapping spheres into the point cloud and evaluate them individually. The final results are generated by a voting scheme. 
In contrast to our method, KPConv~\cite{thomas2019KPConv} performs temporal semantic segmentation by accumulating all clouds of the sequence into one large point cloud and uses no recurrent architecture. 

DarkNet53Seg~\cite{behley2019semantickitti} and TangentConv~\cite{tatarchenko2018tangent} were used as the two baseline networks for the segmentation of 4D point clouds in the SemanticKITTI \cite{behley2019semantickitti} dataset. The input for these were accumulated clouds of the sequences. 
DarkNet53Seg~\cite{behley2019semantickitti} is an extension of SqueezeSeg \cite{wu2018squeezeseg} --- a U-Net architecture with skip connections that uses the spherical projection of LiDAR point clouds to predict a point-wise label map that is refined by a conditional random field and subsequent clustering.
TangentConv~\cite{tatarchenko2018tangent} is based on the notion of tangent convolution --- a different approach to construct convolutional networks on surfaces that assumes that the data is sampled from locally Euclidean surfaces. The input points are projected onto local tangent planes which are used as 2D grids for convolutions. Based on this input, Tatarchenko~\etal\cite{tatarchenko2018tangent} design a U-type network with skip connections. 
In contrast to our approach, both DarkNet53Seg and TangentConv were designed to output dense per-point predictions for single point clouds and contain no recurrent connections. 

\section{Fundamentals}



We use bold upper-case characters to denote matrices and bold lower-case characters to denote vectors. 
We denote with point $p$ a single element from the point cloud and with vertex $v$ an element from the permutohedral lattice. 

The points of a cloud are defined as a tuple $p=\left( \mathbf{g}_p, \mathbf{f}_p \right)$, with $\mathbf{g}_p\in\mathbb{R}^{ d }$ denoting the coordinates of the point and $\mathbf{f}_p \in \mathbb{R}^{ f_d }$ representing the features stored at point $p$ (normals, reflectance, etc.). The full point cloud containing $m$ points is denoted by $P=\left( \mathbf{G}, \mathbf{F} \right)$ with $\mathbf{G}\in\mathbb{R}^{ m \times d }$ denoting the positions matrix and $\mathbf{F}\in\mathbb{R}^{ m \times f_d }$ the feature matrix. 

The vertices of the $d$-dimensional permutohedral lattice \cite{adams2010fast, rosu2019latticenet} are defined as a tuple $v=\left( \mathbf{c}_v, \mathbf{x}_v \right)$, with $\mathbf{c}_v\in\mathbb{Z}^{ (d+1) }$ denoting the coordinates of the vertex and $\mathbf{x}_v \in \mathbb{R}^{ v_d }$ representing the values stored at vertex $v$. A full lattice contains $k$ vertices and is denoted with $V=\left( \mathbf{C}, \mathbf{X} \right)$, with $\mathbf{C}\in\mathbb{Z}^{ k \times (d+1) }$ representing the coordinate matrix and $\mathbf{X}\in\mathbb{R}^{ k \times v_d }$ the value matrix.
For $d=3$ the input space is tessellated into uniform tetrahedra.
We denote the set of neighbors of vertex $v$ with $N(v)$.
The vertices of the permutohedral lattice are stored in a sparse manner using a hash map. 
Hence, we only allocate the simplices that contain the 3D surface of interest. 




Tetrahedras scale linearly in the number of vertices and not quadratically like cubical voxels. This allows for fast interpolation of data from the point cloud to the lattice and backwards.
For further details 
we refer to Rosu~\etal\cite{rosu2019latticenet}.

\section{Architecture}

Our recurrent neural network (RNN) TemporalLatticeNet is an extension of LatticeNet\cite{rosu2019latticenet}. LatticeNet is extended with recurrent connections at multiple resolutions on which temporal information is allowed to flow~(\reffig{fig:rnn__architecture}). 

\subsection{Method}
\label{sec:method}
Input to our network is a sequence of clouds $P = \left(P_0, P_1, ..., P_{n-1}\right)$, where $P_i=\left( \mathbf{G}, \mathbf{F} \right)$ with $n \in \mathbb{N}^+$ and $0 \leq i < n$. We refer to $n$ as the sequence length. The network outputs the likelihood for each possible class for every point $p \in P_n$ (\reffig{fig:overview_arch}). We assume that the points have been transformed in a common reference frame.
The positions $\mathbf{G}$ are scaled by a factor $\pmb{\sigma} \in \mathbb{R}^d$ as $\mathbf{G}_s = \mathbf{G}/\pmb{\sigma}$ which controls the influence area of the permutohedral lattice. If not otherwise stated, we refer to $\mathbf{G}_s$ as $\mathbf{G}$. The matrix $\mathbf{F}$ denoting the per-point features contains the reflectance value from the LiDAR scanner or is set to zeros in the case of a sensor which doesn't output reflectance. 


We insert recurrent connections at various points of the LatticeNet architecture~(\reffig{fig:rnn__architecture}) where the states of two lattices $V^{(t-1)}$ and $V^{(t)}$ have to be fused. 
We refer to the feature matrix of each lattice as $\mathbf{X}^{(t-1)}$ and $\mathbf{X}^{(t)}$, respectively, as state of the network. 
To compute the hidden state $\mathbf{H}^{(t)}$, we fuse the previous hidden state $\mathbf{H}^{(t-1)}$ together with the current state of the network $\mathbf{X}^{(t)}$. For this, a correspondence between the coordinate matrices $\mathbf{C}$ of both lattices has to be known, because they define the order in which the feature vectors $\mathbf{x}_v \in \mathbf{X}$ are saved. This correspondence is achieved by transforming the point clouds into a common frame and using its hash map. By keeping the hash map the same for the whole sequence, the Distribute operation of LatticeNet maps 3D points to the same coordinate vectors $\mathbf{c}$ across timesteps and allows comparing the feature vectors $\mathbf{x}$ with each other. Vertices corresponding to previously unknown areas in the input are inserted at the end of the matrices $\mathbf{C}$ and $\mathbf{X}$ (\reffig{fig:temporal_fuse}).


	
	\bgroup
\newcommand\crule[2][black]{\textcolor{#1}{\rule[-1.5pt]{#2}{#2}}}
	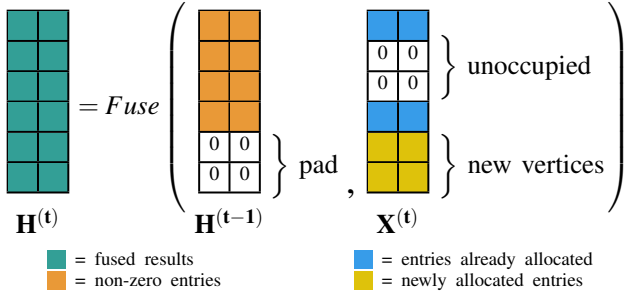
\begin{figure}[]
	
	\resizebox{0.48\textwidth}{!}{
	\begin{tikzpicture}
		\colorlet{featout}{ph-green!80}
		\colorlet{feat1}{ph-orange!80}
		\colorlet{feat2}{ph-blue!80}
		\colorlet{feat3}{ph-green!80}
		\colorlet{feat3}{ph-yellow}
		
		\newcolumntype{M}[1]{>{\centering\arraybackslash}m{#1}}
		\setlength\tabcolsep{0pt}
		\setlength\arrayrulewidth{1pt} 
		
		\node[] at (0,0) {
			\resizebox{1cm}{!}{ 
			\begin{tabular}{|@{\rule[-0.2cm]{0pt}{0.5cm}}*{2}{M{0.5cm} |}}
			\hline
			\cellcolor{featout} & \cellcolor{featout} \\ 
			\hline
			\cellcolor{featout} & \cellcolor{featout} \\ 
			\hline
			\cellcolor{featout} & \cellcolor{featout} \\ 
			\hline
			\cellcolor{featout} & \cellcolor{featout} \\
			\hline
			\cellcolor{featout} & \cellcolor{featout} \\ 
			\hline
			\cellcolor{featout} & \cellcolor{featout} \\  
			\hline
			\end{tabular}
			}
		};
	
		\node[] at (2.5,0) {
			\resizebox{1cm}{!}{ 
			\begin{tabular}{|@{\rule[-0.2cm]{0pt}{0.5cm}}*{2}{M{0.5cm} |}}
			\hline
			\cellcolor{feat1} & \cellcolor{feat1} \\ 
			\hline
			\cellcolor{feat1} & \cellcolor{feat1} \\ 
			\hline
			\cellcolor{feat1} & \cellcolor{feat1} \\ 
			\hline
			\cellcolor{feat1} & \cellcolor{feat1} \\
			\hline
			 0 & 0 \\ 
			\hline
			 0 & 0 \\ 
			\hline 
			\end{tabular}
			}
		}; 
	
		\node[] at (4.7,0) {
			\resizebox{1cm}{!}{  
			\begin{tabular}{|@{\rule[-0.2cm]{0pt}{0.5cm}}*{2}{M{0.5cm} |}}
			\hline
			\cellcolor{feat2} & \cellcolor{feat2} \\ 
			\hline
			0 & 0 \\ 
			\hline
			0 & 0 \\
			\hline
			\cellcolor{feat2} & \cellcolor{feat2} \\
			\hline
			\cellcolor{feat3} & \cellcolor{feat3} \\
			\hline
			\cellcolor{feat3} & \cellcolor{feat3} \\
			\hline  
			\end{tabular}
			}
		};
	
		\node[] at (0.0, -1.55) {
			$\mathbf{H^{(t)}}$
		};
		\node[] at (2.5, -1.55) {
			$\mathbf{H^{(t-1)}}$
		};
		\node[] at (4.7, -1.55) {
			$\mathbf{X^{(t)}}$
		};
	
		\node[] at (7.4, 1.5) {
		};
	
		\node[overlay,anchor=west] at (0.4,0) {			
			$ = Fuse \left(
			\rule{0cm}{1.5cm} \rule{5.5cm}{0cm} \right)
			\rule{0cm}{1.6cm} $ 
		};
	
		\node[overlay,anchor=west] at (2.9, -0.85) {
			\bigg\} pad
		};
		\node[overlay, anchor=west] at (5.1, -0.85) {
			\bigg\} new vertices
		};
		\node[overlay, text width=3cm, anchor=west] at (5.1, 0.45) {
			\bigg\} unoccupied 
		};
	
		\node[overlay, text width=3cm, anchor=west] at (3.9, -1.2) {
			\Large
			,
		};
	
	
	

		\node[text width=3.5cm, anchor=west] at (0.0, -2.2) {
			\scriptsize
			\crule[ph-green!80]{7pt} = fused results\\
			\crule[ph-orange!80]{7pt} = non-zero entries\\
		};
		\node[overlay, text width=3.5cm, anchor=west] at (4.0, -2.2) {
			\scriptsize
			\crule[ph-blue!80]{7pt} = entries already allocated\\
			\crule[ph-yellow]{7pt} = newly allocated entries\\
		};

	\end{tikzpicture}
	}
	
	\caption{Temporal fusion: The feature matrix from the previous time-step $\mathbf{H}^{(t-1)}$ is zero-padded in order to account for the new vertices that were allocated at the current time-step $\mathbf{X}^{(t)}$. The feature matrices are afterwards fused by the chosen recurrent layer. 
} \label{fig:temporal_fuse}
	\end{figure}
	\egroup

\subsection{Position of the Recurrence}
\label{sec:pos_recurrence}

Our RNN is a many-to-one deep RNN whose recurrent layers are positioned at different layers of the network. Four promising positions for the recurrent connections were chosen: Early Fusion, Middle Fusion, Bottleneck Fusion, and Late Fusion as shown in \reffig{fig:rnn__architecture}.   

	\bgroup
\newcommand\crule[2][black]{\textcolor{#1}{\rule[-1.5pt]{#2}{#2}}}
	\def\W{40pt}
	\def\H{10pt}
	\def\Sep{25pt}
	\def\ShiftLevel{20pt}
	\begin{figure}[]
		
		\hspace*{0.07\linewidth}
		\begin{tikzpicture}


			\scriptsize
			
			\def\W{55pt}
			\def\H{12pt}
			\def\Sep{18pt}
			\def\SepLvl{3pt}
			\def\SepUpLvl{5pt}
			\def\ShiftLevel{13pt}
			\def\BrightnessOrange{60}
			\def\BrightnessBlue{60}
			\def\BrightnessGreen{60}
			\def\Brightnessellow{60}
			\colorlet{inputColor}{ph-orange!80}
			\colorlet{distributeColor}{ph-blue!70}
			\colorlet{level1DownColor}{ph-blue!70}
			\colorlet{level2DownColor}{ph-green!70}
			\colorlet{bottleneckColor}{ph-yellow!80}
			\colorlet{level2UpColor}{ph-green!80}
			\colorlet{level1UpColor}{ph-blue!70}
			\colorlet{sliceColor}{ph-blue!70}
			\colorlet{linearColor}{ph-orange!80}
			\colorlet{concatColor}{ph-orange!80}
			\colorlet{temporalFusionColor}{ph-purple!60}
			\def\FSize{\footnotesize}
			\contourlength{0.05em} 
			\contournumber{20}  

		\node[] at (3.4,-3.9) {
			\begin{tikzpicture}[>={Stealth[inset=2pt,length=4.5pt,angle'=28,round]}, scale=0.95, every node/.style={scale=0.95}]

			\node[rectangle,fill=inputColor, text=white, inner sep=0pt, minimum width=\W, minimum height=\H] (input) at (0,+5pt) {\FSize \textbf{Input t}};
			\node[rectangle,fill=level1DownColor, text=white, inner sep=0pt, minimum width=\W, minimum height=\H] (distribute) at (0,-\Sep) {\FSize \textbf{Distribute}};
			\node[rectangle,fill=level1DownColor, text=white, inner sep=0pt, minimum width=\W, minimum height=\H] (pointnet) at (0,-\Sep*2) {\FSize \textbf{PointNet}};
			\node[rectangle,fill=temporalFusionColor, text=white, inner sep=0pt, minimum width=\W, minimum height=\H] (temporalFusionEarly) at (0,-\Sep*3) {\FSize \textbf{Temp. Fusion}};
			
			\node[rectangle,fill=level1DownColor, text=white, inner sep=0pt, minimum width=\W, minimum height=\H] (d1c1) at (0,-\Sep*4) {\FSize \textbf{ResNet Block}};
			\node[rectangle,fill=level1DownColor, text=white, inner sep=0pt, minimum width=\W, minimum height=\H] (d1c2) at (0,-\Sep*5) {\FSize \textbf{ResNet Block}};
			\node[rectangle,fill=temporalFusionColor, text=white, inner sep=0pt, minimum width=\W, minimum height=\H] (temporalFusionMiddle) at (0,-\Sep*6) {\FSize \textbf{Temp. Fusion}};
			
			\node[rectangle,fill=level2DownColor, text=white, inner sep=0pt, minimum width=\W, minimum height=\H] (downsample1) at (0+\ShiftLevel,-\Sep*7-\SepLvl) {\FSize \textbf{Downsample}};
			\node[rectangle,fill=level2DownColor, text=white, inner sep=0pt, minimum width=\W, minimum height=\H] (d2c1) at (0+\ShiftLevel,-\Sep*8-\SepLvl) {\FSize \textbf{ResNet Block}};
			
			\node[rectangle,fill=bottleneckColor, text=white, inner sep=0pt, minimum width=\W, minimum height=\H] (downsample2) at (4+\ShiftLevel*2,-\Sep*9-\SepLvl*2) {\FSize \textbf{Downsample}};
			\node[rectangle,fill=bottleneckColor, text=white, inner sep=0pt, minimum width=\W, minimum height=\H] (bottleneck1) at (4+\ShiftLevel*2,-\Sep*10-\SepLvl*2) {\FSize \textbf{ResNet Block}};	
			\node[rectangle,fill=temporalFusionColor, text=white, inner sep=0pt, minimum width=\W, minimum height=\H] (temporalFusionBottleNeck) at (4+\ShiftLevel*2,-\Sep*11-\SepLvl*2) {\FSize \textbf{Temp. Fusion}};

			\node[rectangle,fill=bottleneckColor, text=white, inner sep=0pt, minimum width=\W, minimum height=\H] (upsample1) at (4+\ShiftLevel*2,-\Sep*12-\SepLvl*2) {\FSize \textbf{Upsample}};
			\node[rectangle,fill=level2UpColor, text=white, inner sep=0pt, minimum width=\W, minimum height=\H] (u1c1) at (0+\ShiftLevel,-\Sep*13-\SepLvl*3-\SepUpLvl) {\FSize \textbf{ResNet Block}};

			\node[rectangle,fill=level2UpColor, text=white, inner sep=0pt, minimum width=\W, minimum height=\H] (upsample2) at (0+\ShiftLevel,-\Sep*14-\SepLvl*3-\SepUpLvl) {\FSize \textbf{Upsample}};
			
			\node[rectangle,fill=temporalFusionColor, text=white, inner sep=0pt, minimum width=\W, minimum height=\H] (temporalFusionLate) at (0,-\Sep*15 -\SepLvl*4-\SepUpLvl*2) {\FSize \textbf{Temp. Fusion}};
			\node[rectangle,fill=level1UpColor, text=white, inner sep=0pt, minimum width=\W, minimum height=\H] (u2c1) at (0,-\Sep*16-\SepLvl*4-\SepUpLvl*2) {\FSize \textbf{ResNet Block}};
			
			\node[rectangle,fill=sliceColor, text=white, inner sep=0pt, minimum width=\W, minimum height=\H] (deform-slice) at (0,-\Sep*17-\SepLvl*4-\SepUpLvl*2) {\FSize \textbf{DeformSlice}};
			\node[rectangle,fill=linearColor, text=white, inner sep=0pt, minimum width=\W, minimum height=\H] (linear) at (0,-\Sep*18-\SepLvl*4-\SepUpLvl*2) {\FSize \textbf{Linear}};
			
			\def\ShiftXConcat{-\W/2 + \ShiftLevel/2} 
			\def\ShiftYConcat{\Sep/2-\H/2+\SepLvl/2+\SepUpLvl/2} 
			\path [draw=black,postaction={on each segment={custom arrow2=black}}, transform canvas={shift={(\ShiftXConcat ,0)}}] (temporalFusionMiddle) to (temporalFusionLate);
			\path [draw=black,postaction={on each segment={mid arrow=black}}, transform canvas={shift={(\ShiftXConcat ,0)}}] (d2c1) to (u1c1);
			
			\node[circle,fill=concatColor, text=white, inner sep=0pt, minimum size=10pt] (concat2) at ([xshift=\ShiftXConcat,yshift=\ShiftYConcat]temporalFusionLate.north) {\footnotesize $\bm{\mathsf{C}}$};
			\node[circle,fill=concatColor, text=white, inner sep=0pt, minimum size=10pt] (concat1) at ([xshift=\ShiftXConcat,yshift=\ShiftYConcat]u1c1.north) {\footnotesize $\bm{\mathsf{C}}$ };
			
			\draw[->] (upsample2) |- (concat2); 
			\draw[->] (upsample1) |- (concat1);
			
			\draw[->] ($(input.south) +(-10pt,0pt) $) --  ($(distribute.north) +(-10pt,0pt) $) node[midway,xshift=2.2pt] (p-middle) {} node[pos=0.5, right] {P} ;
			\draw[->] ($(input.south) +(10pt,0pt) $) -- ($(distribute.north) +(10pt,0pt) $) node[pos=0.5, right] {V};
			\draw[->] (distribute) -- (pointnet);
			\draw[->] (pointnet) -- (temporalFusionEarly); 
			\draw[->] (temporalFusionEarly) -- (d1c1);
			\draw[->] (d1c1) -- (d1c2);
			\draw[->] (d1c2) -- (temporalFusionMiddle);
			\draw[->] (downsample1) -- (d2c1);
			\draw[->] (downsample2) -- (bottleneck1); 	
			\draw[->] (bottleneck1) -- (temporalFusionBottleNeck); 
			\draw[->] (temporalFusionBottleNeck) -- (upsample1); 
			\draw[->] (u1c1) -- (upsample2);
			\draw[->] (temporalFusionLate) -- (u2c1);
			\draw[->] (u2c1) -- (deform-slice);
			\draw[->] (deform-slice) -- (linear);
			
			\path[draw=black,postaction={on each segment={custom arrow={black} }}
			] (p-middle) --  ++(-1,0) |- (deform-slice);
			
			\draw[->] ($ (temporalFusionMiddle.south) + (\ShiftLevel,0) $) -- (downsample1); 
			\draw[->] ($ (d2c1.south) + (4+\ShiftLevel,0) $) -- (downsample2);

			\node[rectangle, text=black, inner sep=0pt, minimum width=\W, minimum height=\H, align = center] (explaintemporalFusionEarly) at (75+\ShiftLevel*0,-\Sep*3) {\FSize Early Fusion};
			
			\node[rectangle, text=black, inner sep=0pt, minimum width=\W, minimum height=\H, align = center] (explaintemporalFusionMiddle) at (75+\ShiftLevel*0,-\Sep*6) {\FSize Middle Fusion};
			
			\node[rectangle, text=black, inner sep=0pt, minimum width=\W+10, minimum height=\H, align = center] (explaintemporalFusionBottleNeck) at (80+\ShiftLevel*0,-\Sep*11-\SepLvl*2) {\FSize Bottleneck\\Fusion};
			
			\node[rectangle,text=black, inner sep=0pt, minimum width=\W, minimum height=\H, align = center] (explaintemporalFusionLate) at (75+\ShiftLevel*0,-\Sep*15 -\SepLvl*4-\SepUpLvl*2) {\FSize Late Fusion};
			
			\end{tikzpicture}
		};

		\node[] at (0,-3.9) {
			\begin{tikzpicture}[>={Stealth[inset=2pt,length=4.5pt,angle'=28,round]}, scale=0.95, every node/.style={scale=0.95}]

			\scriptsize
			
			\node[rectangle,fill=inputColor, text=white, inner sep=0pt, minimum width=\W, minimum height=\H] (input) at (0,+5pt) {\FSize \textbf{Input t-1}};
			\node[rectangle,fill=level1DownColor, text=white, inner sep=0pt, minimum width=\W, minimum height=\H] (distribute) at (0,-\Sep) {\FSize \textbf{Distribute}};
			\node[rectangle,fill=level1DownColor, text=white, inner sep=0pt, minimum width=\W, minimum height=\H] (pointnet) at (0,-\Sep*2) {\FSize \textbf{PointNet}};
			\node[rectangle,fill=white, text=white, inner sep=0pt, minimum width=\W, minimum height=\H] (temporalFusionEarly) at (0,-\Sep*3) {\FSize \textbf{Temp. Fusion}};
			
			\node[rectangle,fill=level1DownColor, text=white, inner sep=0pt, minimum width=\W, minimum height=\H] (d1c1) at (0,-\Sep*4) {\FSize \textbf{ResNet Block}};
			\node[rectangle,fill=level1DownColor, text=white, inner sep=0pt, minimum width=\W, minimum height=\H] (d1c2) at (0,-\Sep*5) {\FSize \textbf{ResNet Block}};
			\node[rectangle,fill=white, text=white, inner sep=0pt, minimum width=\W, minimum height=\H] (temporalFusionMiddle) at (0,-\Sep*6) {\FSize \textbf{Temp. Fusion}};
			
			\node[rectangle,fill=level2DownColor, text=white, inner sep=0pt, minimum width=\W, minimum height=\H] (downsample1) at (0+\ShiftLevel,-\Sep*7-\SepLvl) {\FSize \textbf{Downsample}};
			\node[rectangle,fill=level2DownColor, text=white, inner sep=0pt, minimum width=\W, minimum height=\H] (d2c1) at (0+\ShiftLevel,-\Sep*8-\SepLvl) {\FSize \textbf{ResNet Block}};
			
			\node[rectangle,fill=bottleneckColor, text=white, inner sep=0pt, minimum width=\W, minimum height=\H] (downsample2) at (0+\ShiftLevel*2,-\Sep*9-\SepLvl*2) {\FSize \textbf{Downsample}};
			\node[rectangle,fill=bottleneckColor, text=white, inner sep=0pt, minimum width=\W, minimum height=\H] (bottleneck1) at (0+\ShiftLevel*2,-\Sep*10-\SepLvl*2) {\FSize \textbf{ResNet Block}};	
			\node[rectangle,fill=white, text=white, inner sep=0pt, minimum width=\W, minimum height=\H] (temporalFusionBottleNeck) at (0+\ShiftLevel*2,-\Sep*11-\SepLvl*2) {\FSize \textbf{Temp. Fusion}};

			\node[rectangle,fill=bottleneckColor, text=white, inner sep=0pt, minimum width=\W, minimum height=\H] (upsample1) at (0+\ShiftLevel*2,-\Sep*12-\SepLvl*2) {\FSize \textbf{Upsample}};
			\node[rectangle,fill=level2UpColor, text=white, inner sep=0pt, minimum width=\W, minimum height=\H] (u1c1) at (0+\ShiftLevel,-\Sep*13-\SepLvl*3-\SepUpLvl) {\FSize \textbf{ResNet Block}};
			
			\node[rectangle,fill=level2UpColor, text=white, inner sep=0pt, minimum width=\W, minimum height=\H] (upsample2) at (0+\ShiftLevel,-\Sep*14-\SepLvl*3-\SepUpLvl) {\FSize \textbf{Upsample}};
			
			\node[rectangle,fill=white, text=white, inner sep=0pt, minimum width=\W, minimum height=\H] (temporalFusionLate) at (0,-\Sep*15 -\SepLvl*4-\SepUpLvl*2) {\FSize \textbf{Temp. Fusion}};
			
			\node[rectangle,fill=white, text=white, inner sep=0pt, minimum width=\W, minimum height=\H] (u2c1) at (0,-\Sep*16-\SepLvl*4-\SepUpLvl*2) {\FSize ResNet Block};
			\node[rectangle,fill=white, text=white, inner sep=0pt, minimum width=\W, minimum height=\H] (deform-slice) at (0,-\Sep*17-\SepLvl*4-\SepUpLvl*2) {\FSize DeformSlice};
			\node[rectangle,fill=white, text=white, inner sep=0pt, minimum width=\W, minimum height=\H] (linear) at (0,-\Sep*18-\SepLvl*4-\SepUpLvl*2) {\FSize Linear};
			
			\def\ShiftXConcat{-\W/2 + \ShiftLevel/2} 
			\def\ShiftYConcat{\Sep/2-\H/2+\SepLvl/2+\SepUpLvl/2} 
			\path [draw=black,postaction={on each segment={custom arrow2=black}}, transform canvas={shift={(\ShiftXConcat ,0)}}] (d1c2) to ($ (temporalFusionLate) + (0,0.25) $);
			\path [draw=black,postaction={on each segment={mid arrow=black}}, transform canvas={shift={(\ShiftXConcat ,0)}}] (d2c1) to (u1c1);
			
			\node[circle,fill=concatColor, text=white, inner sep=0pt, minimum size=10pt] (concat2) at ([xshift=\ShiftXConcat,yshift=\ShiftYConcat]temporalFusionLate.north) {\footnotesize $\bm{\mathsf{C}}$};
			\node[circle,fill=concatColor, text=white, inner sep=0pt, minimum size=10pt] (concat1) at ([xshift=\ShiftXConcat,yshift=\ShiftYConcat]u1c1.north) {\footnotesize $\bm{\mathsf{C}}$ };
			
			\draw[->] (upsample2) |- (concat2); 
			\draw[->] (upsample1) |- (concat1);
			
			\draw[->] ($(input.south) +(-10pt,0pt) $) --  ($(distribute.north) +(-10pt,0pt) $) node[midway,xshift=2.2pt] (p-middle) {} node[pos=0.5, right] {P} ;
			\draw[->] ($(input.south) +(10pt,0pt) $) -- ($(distribute.north) +(10pt,0pt) $) node[pos=0.5, right] {V};
			\draw[->] (distribute) -- (pointnet);
			\draw[->] (pointnet) -- (d1c1);
			\draw[->] (d1c1) -- (d1c2);
			\draw[->] (downsample1) -- (d2c1);
			\draw[->] (downsample2) -- (bottleneck1); 
			\draw[->] (bottleneck1) -- (upsample1); 
			\draw[->] (u1c1) -- (upsample2);
			\draw[->] ($ (d1c2.south) + (\ShiftLevel,0) $) -- (downsample1); 
			\draw[->] ($ (d2c1.south) + (\ShiftLevel,0) $) -- (downsample2);

			\fill[fill = white] ($ (pointnet.south) + (0.1,-0.6) $) rectangle ++($ (2.2,0.6) $);
			\draw[overlay,->, -{Stealth[scale=1.0]} , color = moving-car,line width=0.4mm] ($ (pointnet.south) + (0,-0.4) $) -- ($ (pointnet.south) - (-2.2,0.4) $);
			
			\fill[draw = white, fill = white] ($ (d1c2.south) + (1,-0.6) $) rectangle ++($ (1.3,0.6) $);
			\draw[overlay,->, -{Stealth[scale=1.0]} , color = moving-car,line width=0.4mm] ($ (d1c2.south) + (0.48,-0.4) $) -- ($ (d1c2.south) - (-2.2,0.4) $); 
			
			\fill[draw = white, fill = white] ($ (temporalFusionBottleNeck.south) + (0.1,0.0) $) rectangle ++($ (2.3,0.4) $);
			\draw[overlay,->, -{Stealth[scale=1.0]}, color = moving-car,line width=0.4mm] ($ (temporalFusionBottleNeck.south) + (0.0,0.2) $) -- ($ (temporalFusionBottleNeck.south) - (-2.3,-0.2) $);
			
			\fill[draw = white, fill = white] ($ (concat2.south) + (0.1,-0.5) $) rectangle ++($ (2.8,0.5) $);
			\draw[overlay,->, -{Stealth[scale=1.0]}, color = moving-car,line width=0.4mm] ($ (concat2.south) + (0.0,-0.0) $) |- ($ (concat2.south) - (-2.8,0.22) $); 
			
			\end{tikzpicture}
		};
		
	\end{tikzpicture}
	
	\caption{Recurrent architecture: The features from previous time-steps are fused in the current time-step at multiple levels of the network. This allows the network to distinguish dynamic and static objects. Our addition to the LatticeNet architecture are the temporal connections ({\color{moving-car}$\pmb{\rightarrow}$}) and the temporal fusion blocks (\crule[ph-purple!60]{8pt}).}
	\label{fig:rnn__architecture}
	\end{figure}
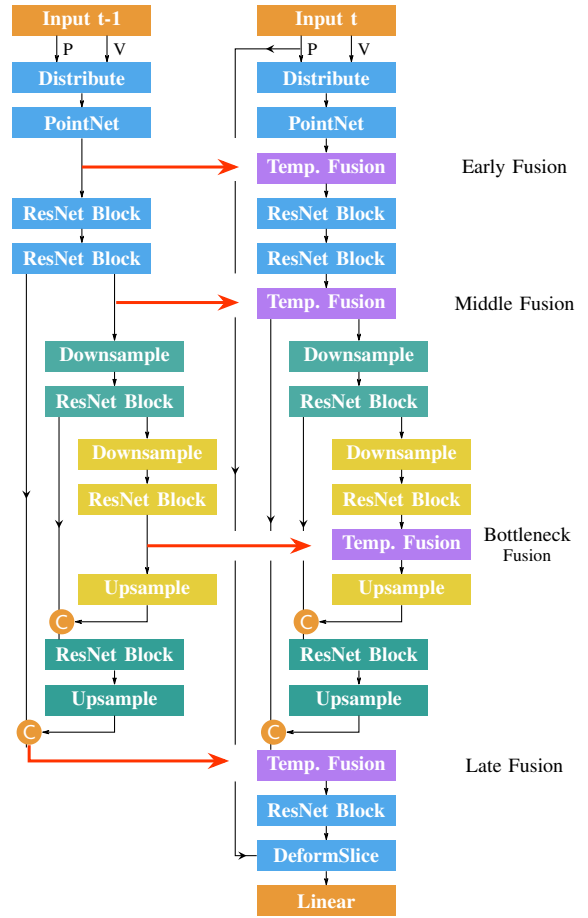
	\egroup

\subsection{Recurrent Layers}
\label{sec:recurrence}

In this section, multiple different recurrent layers are presented and discussed that can be used to fuse $\mathbf{H}^{(t-1)}$ and $\mathbf{X}^{(t)}$. For the last point cloud $P_t$ of the sequence, the previous hidden state $\mathbf{H}^{(n-1)}$ is used to generate the network's prediction. For $t=0$, no computation is performed with $\mathbf{H}^{(0)} = \mathbf{X}^{(0)}$.
In order to fuse $\mathbf{H}^{(t-1)}$ and $\mathbf{X}^{(t)}$, they need the same shape and therefore $\mathbf{H}^{(t-1)}$ is padded with zeros (\reffig{fig:temporal_fuse}).

\textbf{Long short-term memory (LSTM): }
LSTMs \cite{hochreiter1997long} are often used to counteract the problem of vanishing or exploding gradients in sequence learning and show good results for these problems \cite{chung2014empirical}. Therefore we chose them as one recurrent module for our network. 



%

\textbf{Gated recurrent units (GRU): }
As an extension to LSTMs, GRUs were introduced by Cho~\etal\cite{cho2014learning}. They only use an input and a hidden gate and no cell state like LSTMs and are therefore possibly better suited for our task. The performance of GRUs is comparable to LSTMs, while having a lower memory consumption \cite{chung2014empirical}.

\textbf{Abstract Flow (AFlow): }
\begin{figure}[t]
\centering
  \vspace*{3mm}
\includegraphics[width = 0.9\linewidth]{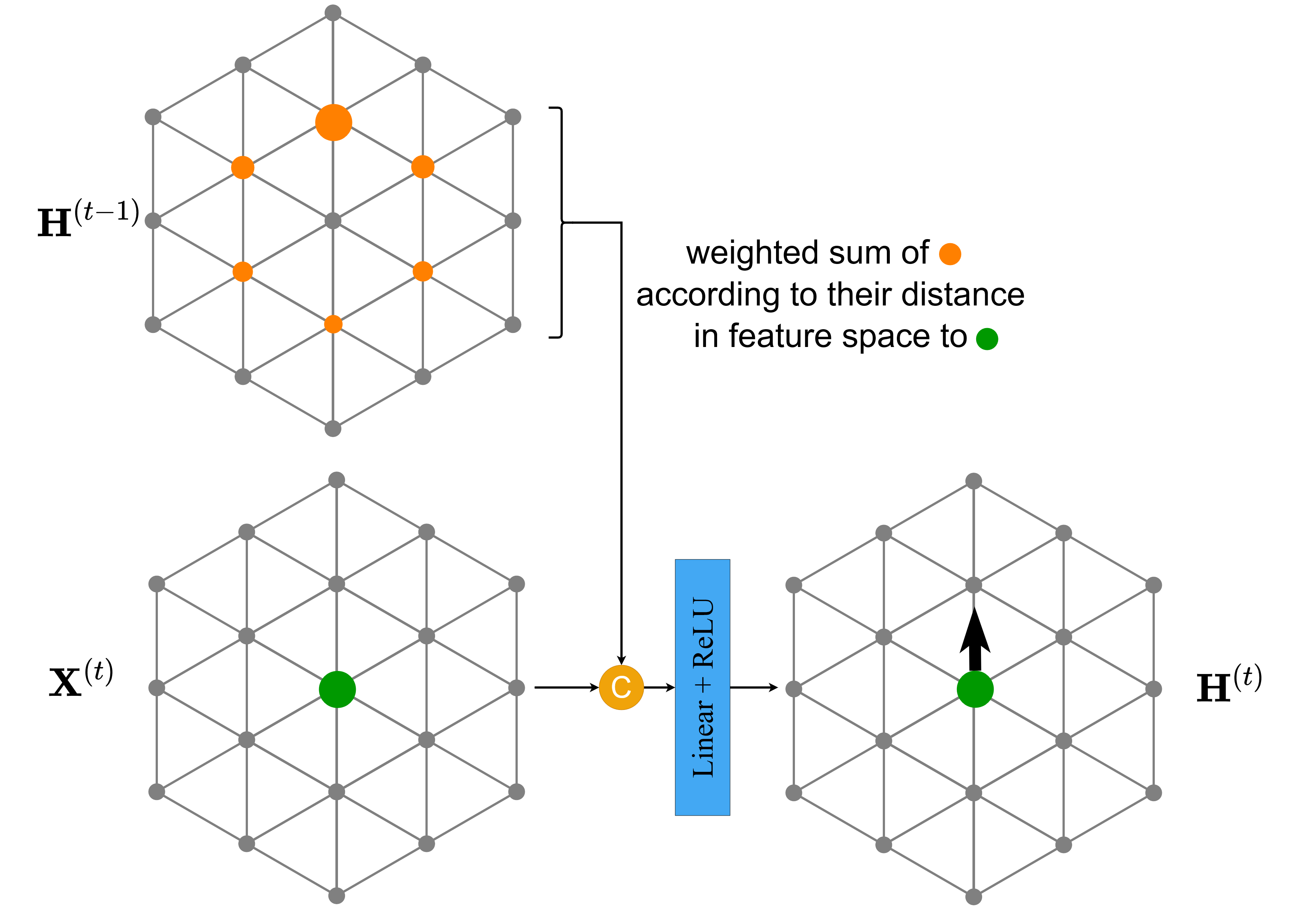}
\caption{
Abstract Flow module: Features from the one-hop neighborhood of the previous timestep $\mathbf{H}^{(t-1)}$ are compared with the center feature $\mathbf{x}_v$ at the current time. A weighted sum is computed based on the distance. The result is concatenated with $\mathbf{x}_v$ and fused using an MLP. A direction can also be established in lattice space between the center feature and the most similar feature from the previous time-step which gives a coarse notion of the movement in the 3D scene as indicated by the arrow. Please note that this figure depicts a 2D example. 
}
\label{fig:cli_module}
\end{figure}
This module is inspired by the CLI module presented by Shi~\etal\cite{shi2020spsequencenet} that aims at combining local information and capturing temporal information between two point clouds.
Our AFlow module can be seen as a convolution with an adaptive convolution kernel. This resembles the ideas presented in Pixel-Adaptive Convolution\cite{su2019pixel}. Therefore, AFlow is designed to extract partial differences between $\mathbf{X}^{(t)}$ and $\mathbf{H}^{(t-1)}$. 
First, the nearest neighbors $N(v)$ of each lattice vertex $v$ with $v \in V^{(t)}$ in the lattice from the previous timestep $V^{(t-1)}$ are found (\reffig{fig:cli_module}). They are used to generate a new local feature vector to fuse temporal information and at the same time summarize the surrounding area from the previous timestep. 
The neighbors $N(v)$ for each lattice vertex are given by the one-hop neighborhood. The neighboring feature vectors from $V^{(t-1)}$ are denoted with $N_{\mathbf{H}}(v)$. For $d=3$, the number of neighbors is given by $|N_{\mathbf{H}}(v)| = 8$.
The feature vectors of the vertices in $N_{\mathbf{H}}(v)$ are weighted according to their distance to the feature vector $\mathbf{x}_v$ in $\mathbf{X}^{(t)}$. The weight is calculated as 
\begin{equation}
\forall  i \in  N_{\mathbf{H}}(v): w_{i} = (\alpha - min(dist(\mathbf{x}_v, \mathbf{h}_i), \alpha))\cdot \beta,
\end{equation}
where $\alpha$ and $\beta$ are learnable parameters that are initialized with $\alpha, \beta = 0.1$. 
Parameter $\alpha$ impacts the maximum distance a neighbor can have from the feature vector we are evaluating at the moment, while $\beta$ controls the maximum value of the resulting feature.  We denote with $dist$ the Euclidean distance between the feature vectors $\mathbf{x}_v$ and  $\mathbf{h}_i$ . 
The AFlow feature matrix $\mathbf{L}$ has the same dimensionality as $\mathbf{X}^{(t)}$. Its feature vectors $\mathbf{l}_v$ are calculated as 
\begin{equation}
\mathbf{l}_v = \sum_{i=1}^{8} \mathbf{h}_i \cdot w_{i}.
\end{equation}
$\mathbf{L}$ is then concatenated with $\mathbf{X}^{(t)}$ along the channels dimension and passed through a linear layer followed by a non-linearity to obtain the new feature matrix $\mathbf{H}^{(t)}$.

The weights calculated in AFlow measure the similarity between features at different timesteps. Similar features that move through time correspond to moving objects in 3D space. We visualize in~\reffig{fig:cli_module} the design of the module and the direction between the center feature and the most similar feature at the previous timestep. Further experiments with the directionality are discussed in \refsec{sec:quali_res}.

\subsection{Network Architectures}
\label{sec:arch_architecture}

The recurrent layers and the recurrence positions can be combined arbitrarily. 
To distinguish the different architectures, the following notation is used: 
The four recurrence layers are separated by a hyphen, e.g. GRU-GRU-AFlow-GRU refers to a network that has a GRU for the early, middle and late fusion and an AFlow module for the bottleneck fusion.



\section{Experiments}

We evaluate our network architectures by calculating the mean Intersection-over-Union (mIoU) on the SemanticKITTI \cite{behley2019semantickitti} dataset. It provides 3D LiDAR-scans from real urban environments and semantic per-point annotations for moving and non-moving classes. We additionally evaluate the impact of the per-point feature matrix $\mathbf{F}$ onto the network: The two possibilities are an empty feature matrix, which forces the network to predict based only on the point positions, and an $\mathbf{F}$ filled with the points' reflectance values.  

\subsection{Implementation}
\label{sec:eval_impl}
All lattice operators with forward and backward passes are implemented on the GPU \cite{rosu2019latticenet} and exposed to PyTorch~\cite{paszke2017automatic}. All convolutions are pre-activated using a ReLU unit \cite{he2016identity,huang2017densely}. For the lattice scale $\pmb{\sigma} = 0.6$ was used and the batch size was chosen as 1.

The models were trained using the Adam optimizer \cite{kingma2014adam, adamw} with a learning rate of $0.001$ and a weight decay of \num{e-4}. 
The learning rate was reduced by a cosine annealing scheduler \cite{sgdr}. The number of epochs between two restarts was chosen as three since less frequent restarts tended to cause over-fitting.
	


\subsection{SemanticKITTI Dataset}
\label{sec:experiments}

The SemanticKITTI \cite{behley2019semantickitti} dataset contains semantically annotated LiDAR scans from the KITTI dataset \cite{Geiger2013IJRR}. The annotations are done for a total of 19 different classes in the single scan task and 25 different classes for the multiple scans task. The scans vary in size from 82K to 129K points with a total of 4,549 million points annotated. In addition to the x, y, and z coordinates, the reflectance value for each point is given. We process each scan entirely without any cropping. 
The training data is augmented with random rotations around the height-axis, mirroring, translation around the other two axes, and random per-point noise. 



\subsection{Generating Predictions} 

The hyperparameters sequence length $n$ and cloud scope $s, \text{with } s \in \mathbb{N}$, have to be chosen for the dataset. The sequence length $n$ defines the cardinality of the input for the network and was chosen as $3 \leq n \leq 5$. We observed that $n < 3$ doesn't allow the network to aggregate enough information, while $n > 5$ leads to memory and time constraints. A sequence length of $n = 4$ worked best for our models. 
For SemanticKITTI $s = 3$ was chosen, which means that between clouds in the sequence $P$ two clouds in the dataset are skipped. 
We've found that using directly consecutive clouds ($s = 1$) can negatively impact the segmentation results since the input clouds are too similar and therefore choose for all our experiments a cloud scope $s=3$.

It is to be noted that during training we need to keep all input clouds in memory in order to perform back-propagation through time, while during inference we evaluate only the cloud at the current time-step since the features from previous time-step are already stored in memory. This recursive formulation yields inference speed that is similar to the original LatticeNet architecture which operated on a single-scan.



\subsection{Quantitative Results}
\label{sec:quant_res}

\bgroup
\def\checkmark{\tikz\fill[scale=0.2](0,.35) -- (.25,0) -- (1,.7) -- (.25,.15) -- cycle;} 
\newcolumntype{Q}{C{2cm}}
\begin{table}
    \vspace*{1mm}
    \caption{Results on SemanticKITTI for different versions of our own architecture. The network with the best result is used in \reftab{tab:semanticKitti_comparison} for comparisons.}
    \centering
    \begin{tabularx}{\columnwidth}{L{4cm}c|Q}
        \toprule 
        Approach & mIoU &with reflectance\\
        \midrule
        LSTM-LSTM-AFlow-LSTM   & 46.7 & \checkmark \\
        GRU-GRU-AFlow-AFlow   & 46.9 & \checkmark \\
        GRU-GRU-AFlow-GRU   & \textbf{47.1} & \checkmark\\
        GRU-GRU-/-GRU  & 44.1 & \checkmark\\
        \midrule
        GRU-GRU-AFlow-GRU  & 42.8 & x\\
        LatticeNet-MLP~\cite{rosu2021latticenetauro}  & 45.2 & x\\
        \bottomrule
    \end{tabularx}    
    \label{tab:semanticKitti_comparison_inside}
\end{table}
\egroup

\bgroup
\newcolumntype{Q}{C{0.19cm}}
\begin{table*}
    \scriptsize
    \vspace*{1mm}
    \caption{State-of-the-art results on SemanticKITTI in comparison to our best performing network.\protect\footnotemark}
    \centering
    \begin{tabularx}{\textwidth}{L{2.2cm}c|QQQQQQQQQQQQQQQQQQQQQQQQQ}
        \toprule 
        Approach & \begin{sideways}mIoU\end{sideways} & \begin{sideways}car\end{sideways} & \begin{sideways}truck\end{sideways} & \begin{sideways}other-vehicle\end{sideways} & \begin{sideways}person\end{sideways} & \begin{sideways}bicyclist\end{sideways} & \begin{sideways}motorcyclist\end{sideways} & \begin{sideways}road\end{sideways} & \begin{sideways}parking\end{sideways} & \begin{sideways}sidewalk\end{sideways} & \begin{sideways}other-ground\end{sideways} & \begin{sideways}building\end{sideways} & \begin{sideways}fence\end{sideways} & \begin{sideways}vegetation\end{sideways} & \begin{sideways}trunk\end{sideways} & \begin{sideways}terrain\end{sideways} & \begin{sideways}pole\end{sideways}& \begin{sideways}traffic sign\end{sideways}& \begin{sideways}moving-car\end{sideways}& \begin{sideways}moving-bicyclist\end{sideways}& \begin{sideways}moving-person\end{sideways}& \begin{sideways}moving-motorcyclist\end{sideways}& \begin{sideways}moving-other-vehicle\end{sideways}& \begin{sideways}moving-truck\end{sideways}\\
        \midrule
        TangentConv\cite{tatarchenko2018tangent}  & 34.1 & 84.9	&2.0	&18.2	&21.1	&18.5	&1.6		&83.9	&38.3	&64.0	&15.3	&85.8	&49.1	&79.5	&43.2	&56.7	&36.4	&31.2	&40.3	&1.1	&6.4	&1.9	&\textbf{30.1}	&\textbf{42.2} \\
        DarkNet53Seg\cite{behley2019semantickitti}  & 41.6 & 84.1	&30.4	&32.9	&20.2	&20.7	&7.5		&91.6	&\textbf{64.9}&75.3	&\textbf{27.5}	&85.2	&56.5	&78.4	&50.7	&64.8	&38.1	&53.3	&61.5	&14.1	&15.2	&0.2	&28.9	&37.8 \\
        SpSequenceNet\cite{shi2020spsequencenet}   & 43.1 & 88.5	&24.0	&26.2	&29.2	&22.7	&6.3		&90.1	&57.6	&73.9	&27.1	&91.2	&\textbf{66.8}	&84.0	&66.0	&65.7	&50.8	&48.7	&53.2	&41.2	&26.2	&36.2	&2.3	&0.1  \\ 
        KPConv\cite{thomas2019KPConv}  & \textbf{51.2} & \textbf{93.7}& 44.9& \textbf{47.2}& \textbf{42.5}& \textbf{38.6}& \textbf{21.6}& 86.5& 58.4& 70.5& 26.7& 90.8& 64.5& \textbf{84.6}& \textbf{70.3}& 66.0& 57.0& 53.9& \textbf{69.4}& \textbf{67.4}& \textbf{67.5}& 47.2& 4.7& 5.8 \\ 
        TemporalLidarSeg\cite{duerr2020lidar}  & 47.0 & 92.1	&\textbf{47.7}&40.9	&39.2	&35.0	&14.4		&\textbf{91.8}&59.6	&\textbf{75.8}	&23.2	&89.8	&63.8	&82.3	&62.5	&64.7	&52.6	&\textbf{60.4}	&68.2	&42.8	&40.4	&12.9	&12.4	&2.1 \\ 
        LatticeNet-MLP\cite{rosu2021latticenetauro} & 45.2 & 91.1& 16.8& 25.0& 29.7& 23.1& 6.8& 89.7& 60.5& 72.5& 26.9& \textbf{91.9}& 64.7& 82.9& 65.0& 63.7& 54.7& 47.1& 54.8& 44.6& 49.9& \textbf{64.3}& 0.6& 3.5 \\
        \midrule
        Ours & 47.1 & 91.6& 35.4& 36.1& 26.9& 23.0& 9.4& 91.5& 59.3& 75.3& \textbf{27.5}& 89.6& 65.3& \textbf{84.6}& 66.7& \textbf{70.4}& \textbf{57.2}& \textbf{60.4}& 59.7& 41.7& 9.4& 48.8& 5.9& 0.0 \\
        \bottomrule
    \end{tabularx}
    \label{tab:semanticKitti_comparison}
\end{table*}
\egroup

We evaluated five different variants of our architectures on the test set of SemanticKITTI and report their resulting mIoU in \reftab{tab:semanticKitti_comparison_inside}. 
We observe that the AFlow module resulted in an improved mIoU of 3.0 points, compared to the base-network that only utilizes GRUs. Adding another AFlow layer at the late fusion instead of an GRU resulted in slightly worse results, which can be explained by the lower feature dimension and therefore a lack of comparability in feature space.

Without using the reflectance as input, the result deteriorated significantly. 
The reason for this might be that the reflectance is a very useful feature for distinguishing similar feature vectors.
This applies to the other recurrent blocks as well, albeit not as severely.
Additionally, we evaluate the performance of the original LatticeNet~\cite{rosu2019latticenet} on temporal data. For this experiment, we accumulated all clouds of a sequence and used this cloud as input. The results were worse than all comparable temporal networks. 

To compare our network's results to the state-of-the-art on SemanticKITTI, we chose the best performing network GRU-GRU-AFlow-GRU. The IoU for all 25 classes are presented in \reftab{tab:semanticKitti_comparison}. We improved performance in relation to our previously published architecture LatticeNet-MLP~\cite{rosu2019latticenet}. 

Our best network has outstanding performance in the segmentation of the classes ‘vegetation’, ‘terrain’, ‘pole’, and ‘traffic sign’. An explanation for the results on ‘traffic sign’ and ‘pole’ is that they are only represented by a small number of points and having multiple clouds can help with a better segmentation quality.

Our network's performance is comparable to TemporalLidarSeg~\cite{duerr2020lidar}, but is outperformed by KPConv~\cite{thomas2019KPConv} with a mIoU that is smaller by $4.1$ points. It is to be noted that KPConv~\cite{thomas2019KPConv} cannot process the whole cloud due to memory constraints, but has to rely on fitting multiple spheres into the cloud to ensure that each point is tested multiple times. The final result is then determined by a voting scheme, in contrast to our approach that processes the whole cloud at once with a single prediction.
TemporalLidarSeg~\cite{duerr2020lidar}, on the other hand, relies on the spherical projection of the 3D cloud to perform 2D operations, while our approach is able to utilize the 3D cloud without any projection.


\bgroup
\newcolumntype{M}{L{4cm}}
\newcolumntype{Q}{R{1.0cm}}
\newcolumntype{P}{L{1.0cm}}
    \begin{table}
    \footnotesize
    \caption{Average time used by the forward pass and the maximum memory used during training. }
    \centering
    \begin{tabularx}{\columnwidth}{M|QP}
        \toprule
                    & \multicolumn{2}{c}{SemanticKITTI}  \\ 
                    & \multicolumn{1}{r}{$\left[\SI{}{\milli\second}\right]$} & \multicolumn{1}{l}{$\left[\SI{}{\giga\byte}\right]$}\\
                    \midrule
        
        LSTM-LSTM-AFlow-LSTM & 151\ph&\ph20 \\
        GRU-GRU-AFlow-GRU & 154\ph&\ph20 \\
        GRU-GRU-AFlow-AFlow & 159\ph&\ph22 \\
        GRU-GRU-/-GRU & 140\ph&\ph18 \\			
        \midrule
        KPConv\cite{thomas2019KPConv}  & 225\ph&\ph15 \\
        SpSequenceNet\cite{shi2020spsequencenet}  & 477\ph&\ph3 \\
        \bottomrule
    \end{tabularx}
    \label{tab:perf-results}
\end{table}
\egroup

We present the performance results in \reftab{tab:perf-results}. The measurements were taken on a NVIDIA GeForce RTX 3090 and the inference time was measured on the validation set.  Each AFlow module increases the inference time and memory consumption, caused by the high number of weights in the AFlow module and the distance calculation per vertex. We are able to segment the cloud faster than KPConv\cite{thomas2019KPConv}, because we are able to reuse feature matrices from previous segmentations due to our recurrent architecture. 


\subsection{Qualitative Results}
\label{sec:quali_res}

\bgroup
\newcommand\crule[2][black]{\textcolor{#1}{\rule[-1.5pt]{#2}{#2}}}
\begin{figure}[t]
\centering
  \vspace*{3mm}
\includegraphics[width = \linewidth]{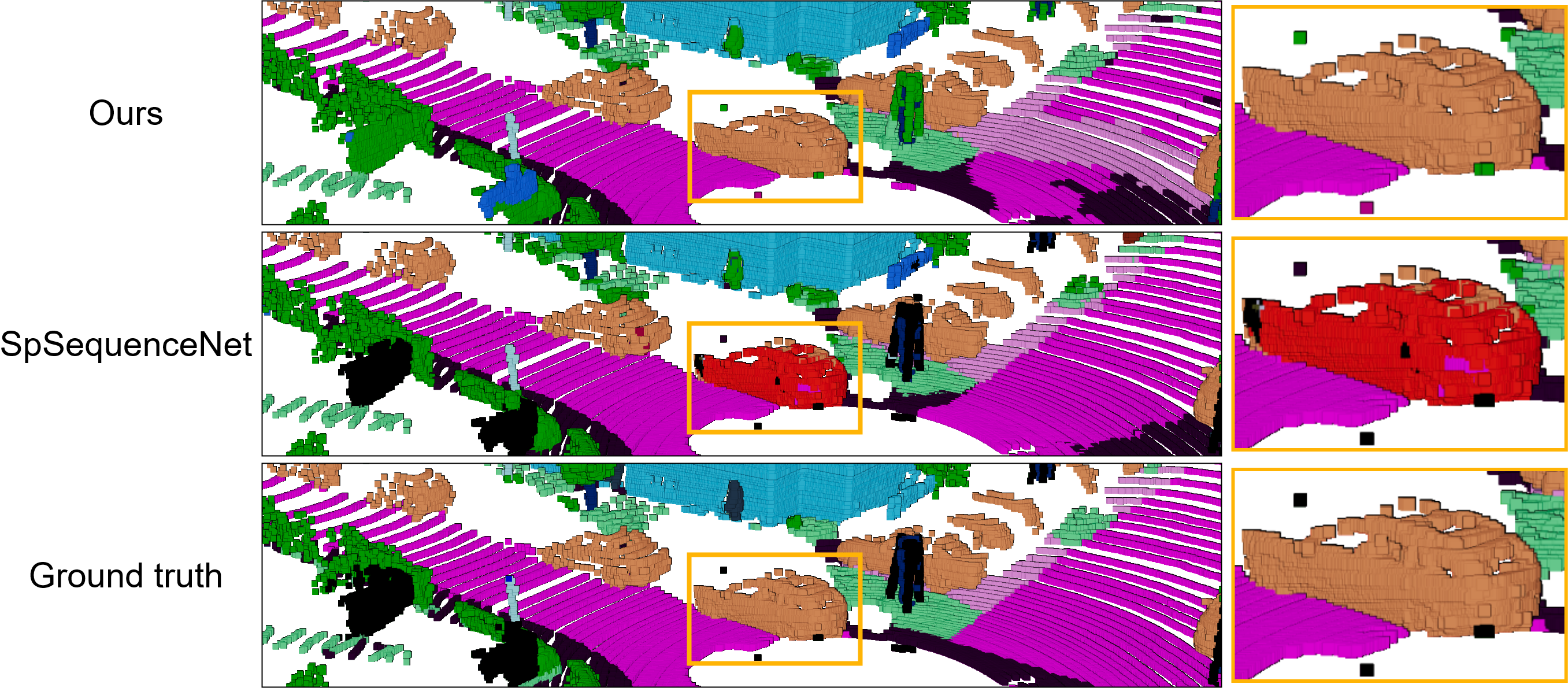}
\caption{In comparison to SpSequenceNet, we are able to better segment stationary~(\crule[car]{8pt}) and moving cars~(\crule[moving-car]{8pt}) in small streets with a high number of cars in the vicinity. SpSequenceNet \cite{shi2020spsequencenet} was the best network that provided pretrained models and was therefore chosen as the comparison. In this example, it is able to better distinguish the road~(\crule[road]{8pt}) from a sidewalk~(\crule[parking]{8pt}). 
Unlabeled~(\crule[unlabeled]{8pt}) points are ignored during training.
}
\label{fig:comparison_images}
\end{figure}
\egroup
 
We present a visual comparison of the segmentation quality of our method with SpSequenceNet in \reffig{fig:comparison_images}.

\bgroup
\newcommand\crule[2][black]{\textcolor{#1}{\rule[-1.5pt]{#2}{#2}}}
\begin{figure}
\centering
\includegraphics[width = \linewidth]{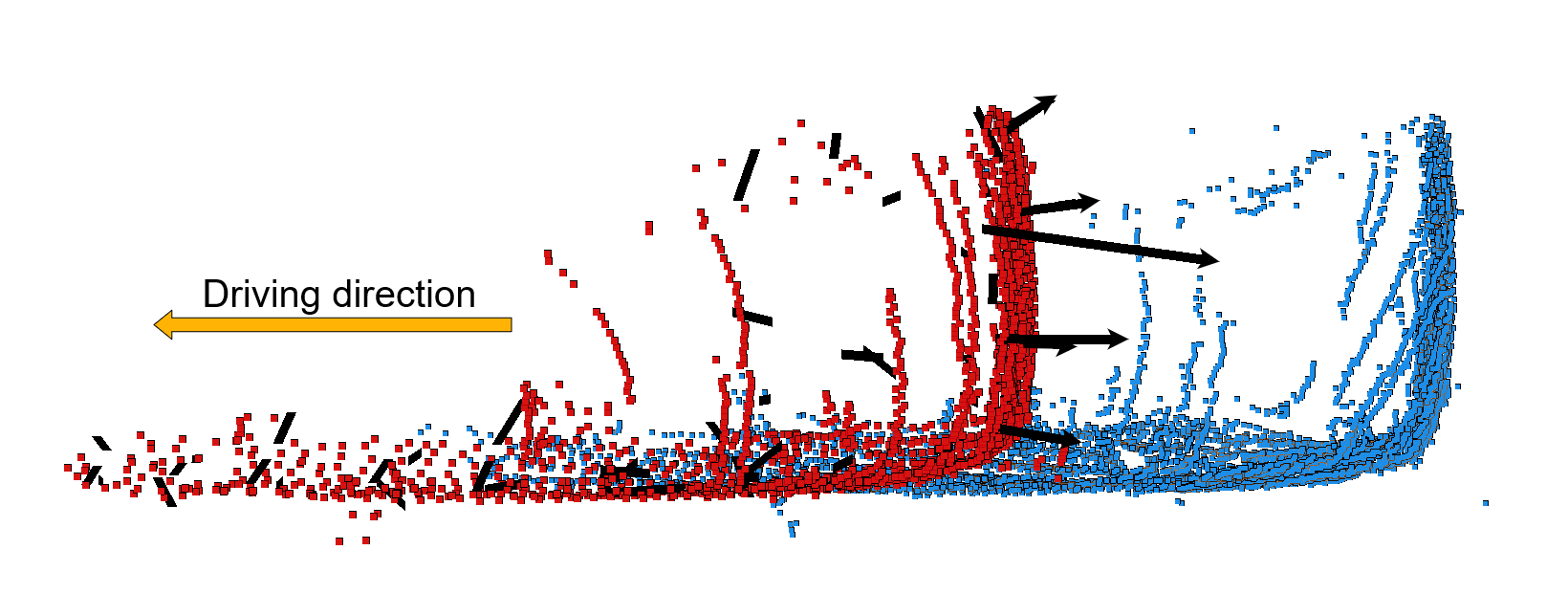}
\caption{Visualization of the AFlow module on the segmentation: Birds-eye view of a car at two different time-steps. The correspondence between the car in the previous timestep~(\crule[ph-dark-blue!80]{8pt}) and the current timestep~(\crule[ph-dark-red]{8pt}) is made by the module and therefore the car is correctly segmented as moving-car.}
\label{fig:cli}
\end{figure}
\egroup

In order to analyze the effects of the AFlow model, we mapped the previously mentioned directionality to 3D space in order to show a coarse direction of movement of the 3D objects. Lattice vertices are approximated in 3D by the average of the 3D points that contribute to them. In~\reffig{fig:cli}, we show one car at two different timesteps. For each lattice vertex in 3D, we draw an arrow showing the direction of the most similar feature from the current time towards the previous timestep. We see that for a car driving towards the left, the directionality from AFlow corresponds to the inverse of the driving direction.



\bgroup
\newcommand\crule[2][black]{\textcolor{#1}{\rule[-1.5pt]{#2}{#2}}}
\begin{figure}
\centering
\begin{subfigure}[b]{\linewidth}
   \includegraphics[width=\linewidth]{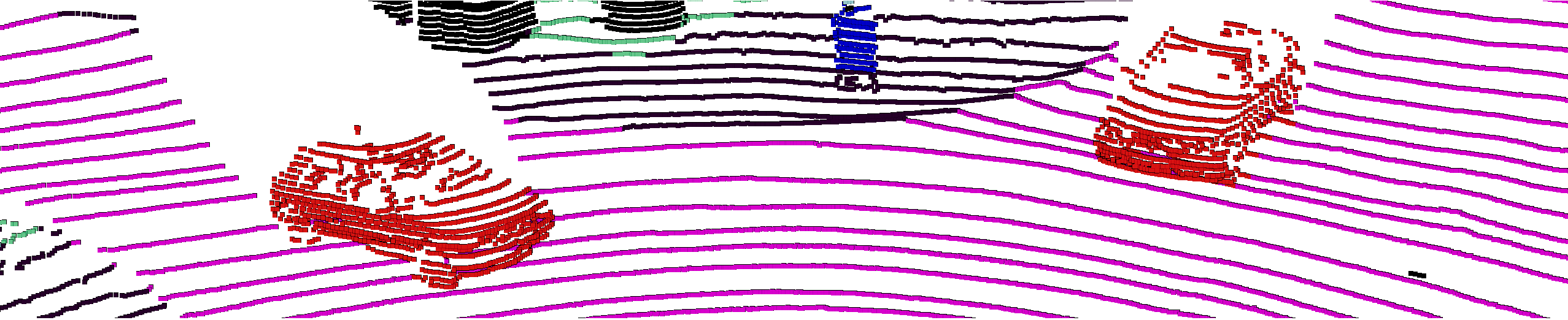}
	\vspace*{-6mm}
   \caption{Ground truth segmentation.}
   \label{fig:gt_failure} 
\end{subfigure}

\vspace*{3mm}
\begin{subfigure}[b]{\linewidth}
   \includegraphics[width=\linewidth]{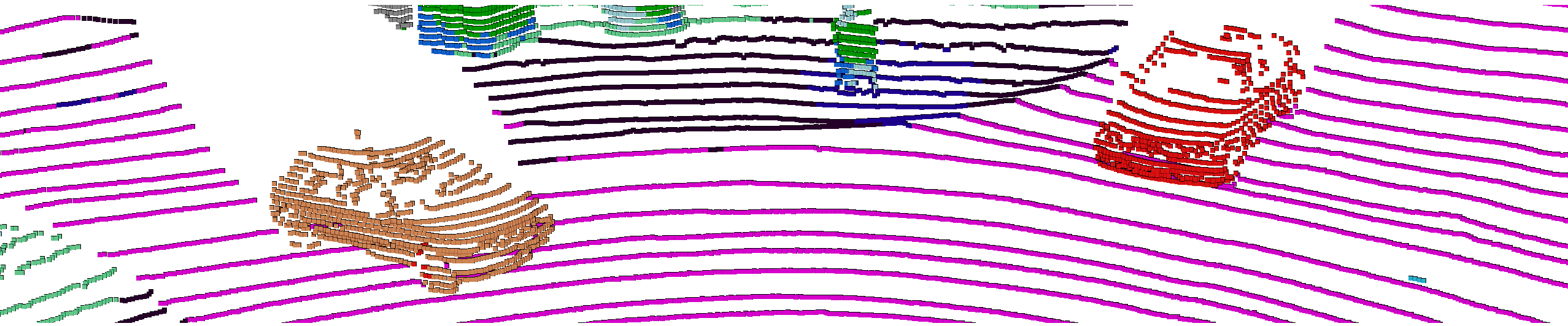}
   \vspace*{-6mm}
   \caption{Predicted segmentation.}
	
   \label{fig:failure}
\end{subfigure}
\caption{Failure case: The prediction fails for the car on the left side, because it is predicted as car~(\crule[car]{8pt}) instead of moving-car~(\crule[moving-car]{8pt}). The reason for this is that the car is waiting at the crossroads for many timesteps.}
\label{fig:failure_prediction}
\end{figure}
\egroup

A failure case of our architecture is moving objects that are waiting/standing still for a duration that exceeds our temporal scope. An example of a car that is waiting at a crossroad is presented in \reffig{fig:failure}. This shouldn't result in problems for an autonomous agent that takes actions based on this segmentation, because the object actually is standing still and is correctly classified as moving once it starts driving again. 

\section{Conclusion}

\footnotetext{We only compare to already published approaches.}
We presented an extension to the original LatticeNet~\cite{rosu2019latticenet} for temporal semantic segmentation. We evaluate different recurrence modules and propose a novel Abstract Flow module that better integrates temporal information. On the SemanticKITTI dataset we achieve comparative results with other baselines while running faster and being able to process the full point cloud at once.

\addtolength{\textheight}{-12cm}   




\section*{Acknowledgment}
This work has been funded by the German Federal Ministry of Education and Research (BMBF) in the project ”Kompetenzzentrum: Aufbau des Deutschen Rettungsrobotik-Zentrums” (A-DRZ).


\clearpage
\bibliographystyle{IEEEtran}
\bibliography{references}

\end{document}